\documentclass[conference,10pt]{IEEEtran}
\usepackage{epsfig,rotating,setspace,latexsym,amsmath,epsf,amssymb,amsfonts,bm,theorem,subfigure,epstopdf}
\usepackage{cite,authblk}
\usepackage{bbm}
\usepackage{color}
\usepackage{mathtools}
\usepackage{algorithm}
\usepackage{algpseudocode}
\usepackage{verbatim}
\usepackage{xcolor}

\algrenewcommand\algorithmicforall{\textbf{foreach}}
\algrenewcommand\algorithmicindent{.8em}

\algnewcommand\algorithmicforeach{\textbf{for each}}
\algdef{S}[FOR]{ForEach}[1]{\algorithmicforeach\ #1\ \algorithmicdo}

\IEEEoverridecommandlockouts
\allowdisplaybreaks

\begin{document}

\title{RAG-Check: Evaluating Multimodal Retrieval Augmented Generation Performance}

\author{
Matin Mortaheb$^{\dag}$, Mohammad A. (Amir) Khojastepour$^{*}$, Srimat T. Chakradhar$^{*}$, Sennur Ulukus$^{\dag}$ \\
\normalsize $^{\dag}$University of Maryland, College Park, MD, $^{*}$NEC Laboratories America, Princeton, NJ \\
\normalsize \emph{mortaheb@umd.edu, amir@nec-labs.com, chak@nec-labs.com, ulukus@umd.edu}
}

\maketitle

\begin{abstract}
Retrieval-augmented generation (RAG) improves large language models (LLMs) by using external knowledge to guide response generation, reducing hallucinations. However, RAG, particularly multi-modal RAG, can introduce new hallucination sources: (i) the retrieval process may select irrelevant pieces (e.g., documents, images) as \underline{raw context} from the database, and (ii) retrieved images are processed into \underline{text-based context} via vision-language models (VLMs) or directly used by multi-modal language models (MLLMs) like GPT-4o, which may hallucinate. To address this, we propose a novel framework to evaluate the reliability of multi-modal RAG using two performance measures: (i) the relevancy score (RS), assessing the relevance of retrieved entries to the query, and (ii) the correctness score (CS), evaluating the accuracy of the generated response. We train RS and CS models using a ChatGPT-derived database and human evaluator samples. Results show that both models achieve ~88\% accuracy on test data. Additionally, we construct a 5000-sample human-annotated database evaluating the relevancy of retrieved pieces and the correctness of response statements. Our RS model aligns with human preferences 20\% more often than CLIP in retrieval, and our CS model matches human preferences ~91\% of the time. Finally, we assess various RAG systems' selection and generation performances using RS and CS.
\end{abstract}

%

\section{Introduction}
\label{sec:intro}

Generative models have seen significant improvements with recent advancements in large language models (LLMs) \cite{achiam2023gpt}. Although the generated responses often reach human-like quality, hallucinations—generating incorrect or irrelevant responses—remain an issue \cite{ji2023survey}. This problem is particularly concerning for applications where accuracy is critical, such as in medical evaluations, processing insurance claims, and autonomous decision-making.
The hallucination issue also persists in vision-language models (VLMs) \cite{liu2024visual, InstructBLIP, lin2024vila, team2023gemini}, which process both images and user queries to generate text responses. Several robust VLMs, such as LLaVA \cite{liu2024visual}, InstructBLIP \cite{InstructBLIP}, and VILA \cite{lin2024vila}, exist. However, these models sometimes produce incorrect responses based on the provided images and user queries.

\begin{figure}[t]
\centerline{\includegraphics[width=1\linewidth]{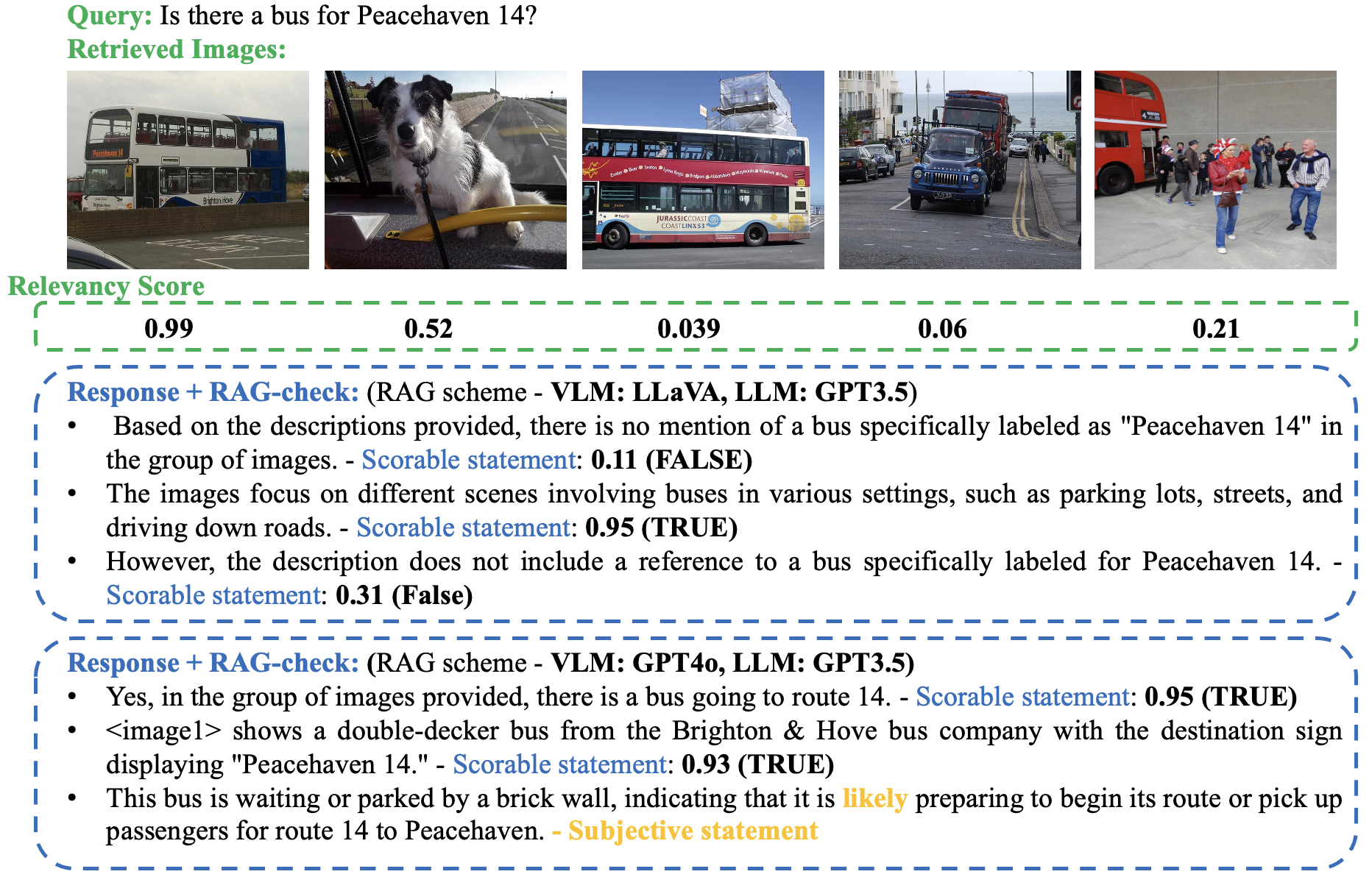}}
  \caption{RAG-check example.}
\centering
\label{fig:RAG_example}
\vspace*{-0.5cm}
\end{figure}

On the other hand, retrieval-augmented generation (RAG) \cite{lewis2020retrieval} offers a promising solution to improve the relevance of responses generated by LLMs. RAG systems enhance LLMs by incorporating external knowledge, on which the user seeks to base their responses. In a RAG system, external knowledge such as enterprise data is stored in a database. When a user submits a query, the RAG system retrieves a few relevant and similar pieces of data from the database. The LLM then generates a coherent response based on this curated information. This approach reduces hallucinations by constraining the LLM to generate responses grounded in the provided external knowledge, thereby increasing accuracy and relevance.
However, in addition to the possible hallucinations in the response, the RAG scheme introduces some new sources of hallucinations during the retrieval process including the selection of the information pieces and the context generation. During the retrieval process, depending on the selection algorithm, RAG selects a few, say $k$, entries from the database that have the highest similarity to the user query. A common approach is to select the $k$ entries whose embeddings have the highest cosine similarity to the query embedding, referred to as the top-$k$ entries. However, this method of selection may not always retrieve the most relevant data for the user query. The wrong selection can result in an incorrect response since the LLM would never see the correct information. We refer to this as \emph{selection-hallucination}. Another source of hallucination is during context generation in the RAG scheme. 
In particular, multi-modal RAG systems process each selected piece of information to text and generate a text-based context by concatenating the text-based context for all the selected pieces. The text-based context is then provided to the LLM in the final stage of the RAG system along with the query to produce the response. For example, if the selected piece is an image, VLM may be used, say with a prompt ``Describe the image" (DTI) to narrate the images into the text, and it is well-known that VLMs may hallucinate and produce wrong context in this case. We refer to this as \emph{context-generation-hallucination}. In the final stage, LLM may also hallucinate when producing the response which is referred to as \emph{response-generation-hallucination}.

Prior works \cite{min2023factscore, chuang2024lookback, bakman2024mars} provide means to evaluate LLMs and measure the correctness of text-based responses by LLM given a text query. FactScore \cite{min2023factscore} demonstrates that breaking long statements into fine-grained atomic statements can improve the identification of hallucinations. This method allows for more precise verification of each individual statement against the source material.
Lookback Lens \cite{chuang2024lookback} detects context hallucination by analyzing the attention scores of the corresponding LLM. It calculates a ratio attention scores between query tokens and response tokens to identify parts of the response that may not be adequately supported by the context. MARS \cite{bakman2024mars} builds on the idea that some parts of a statement are more crucial than others in determining hallucination. The work highlights such important part by assigning more weight to these significant parts, thereby improving the detection of inaccuracies by focusing on the most relevant segments. Despite these advancements, a few papers have specifically addressed hallucination detection in text-based RAG. Among these, RAGAS \cite{es2023ragas} and LlamaIndex evaluation stand out. RAGAS focuses on evaluating the accuracy and relevance of responses generated by RAG systems. It assesses how well the retrieved documents support the generated response and whether the information is correctly incorporated. RAGAS typically involves human evaluators who rate the coherence and factual accuracy of the responses. LlamaIndex relies on GPT4 \cite{achiam2023gpt} to evaluate the faithfulness of the generated response and the relevancy of retrieved documents. Note that both RAGAS and LlamaIndex are designed for text-based queries and datasets.

On the other hand, some works have focused on detecting hallucination in VLMs \cite{gunjal2024detecting, jing2023faithscore, jiang2024hal}. The work in \cite{gunjal2024detecting} presents a model inspired by InstructBLIP to detect hallucinations occurring in single-image scenarios with user queries. This approach leverages the capabilities of InstructBLIP to scrutinize the generated responses for inaccuracies related to the provided image and query. FaithScore \cite{jing2023faithscore} is another tool trained to identify the correctness of VLM-generated statements with an image input. FaithScore first extracts a comprehensive list of atomic facts from the generated response, then verifies the accuracy of these fine-grained atomic facts against the input image.
However, to the best of our knowledge, there are no existing works that provide hallucination scores specifically for multi-modal RAG systems, where the contexts include multiple pieces of multi-modal data (say multiple images) which is retrieved from a database and then the inference is performed on such retrieved multi-modal context. Multi-modal RAG systems present unique challenges, as they require integrating information from various visual sources along with the text to generate coherent and accurate responses. 

Our goal is to develop RAG-check, a method to evaluate the performance of multi-modal RAG schemes. 
RAG-check comprises three main components: (i) First, we design and train a neural network structure that takes the selected pieces of data by the RAG scheme by tapping into the internal components of a RAG system and producing a relevancy score (RS) between each selected piece and the query. For example, when an image (or text) is selected, RS evaluates how relevant the retrieved image (or text) is to the user query. By assigning a relevancy score to each image, we can determine how well the visual data aligns with the user's intent and the specific information requested. (ii) Second, we design an algorithm to partition the output response by the RAG into segments, namely \emph{spans}, and categorize each span. The \emph{span} may be defined as statements, phrases, etc. Some spans may not be scorable, e.g., if it is based on personal opinions or feelings rather than on facts. Alternatively, a span is not scorable if it states an analysis say in a conditional statement, or expresses the uncertainty such as the possibility or probability of something being true. We label such statements as ``subjective" which have been also referred to as ``analysis" statement type in the prior-art \cite{gunjal2024detecting}. A span that is not ``subjective", i.e., it is scorable, is labeled as ``objective".
(iii) Third, we design and train a neural network structure which assesses the correctness of each objective span in the view of the \emph{raw context} defined as the selected pieces of the data by the RAG. We note that in multi-modal RAG, the raw context may be first converted to \emph{text-based context}, e.g., by using VLM for the images, and then text-based context will be used by an LLM to generate the response based on the query. Alternatively, the raw context may be directly fed to a MLLM to generate the response based on the query. Irrespective of the internal structure of the RAG system, the CS assesses the accuracy of each span of the generated text with the raw context which include the original set of the selected pieces. Indeed, the CS measures how correctly each part of the response reflects the information presented in the raw context. By evaluating the correctness of the text spans, we ensure that the generated responses are not only relevant but also factually accurate based on the multi-modal data. 

The contributions of this paper are as follows: (i) We design and train a neural network structure to find the RS between the query and each selected piece of data by RAG scheme.
(ii) We perform partitioning the RAG response into spans and categorizing each span as ``subjective" or ``objective". The partitioning and categorization is performed based on developing a dedicated algorithm as well as using GPT3.5 API. 
(iii) We design and train a neural network structure to find the CS between the raw context, i.e., the selected pieces of information from the database by RAG, and each objective span of the generated response by RAG.
(iv) In addition to the original training, validation, and testing dataset used for developing CS and RS, we also build a database containing 5000 samples by gathering human feedback (HF) on the RAG selection process and the accuracy of the RAG output spans. This collected HF was then used to assess the performance of the designed RS and CS scores with human evaluation for a RAG system.
(v) While CS and RS can be used to evaluate the reliability of an instance of the response generated by RAG, we show a different use of the tools developed in this paper by comparing the average performance of different RAG systems in terms of the relevancy of their retrieval and the correctness of their responses.

\section{Multimodal RAG}
A RAG system is composed of a database, namely, enterprise database, which is usually pre-processed by partitioning the data into pieces and generating an embedding for each piece. The use of embeddings serves two main purposes: (i) To make a compact representation of each piece of information in order to enhance the retrieval process. 
(ii) To have speedy retrieval usually by using a similarity search using dot product (i.e., cosine similarity). 

In a multi-modal RAG, depending on the type of data, embeddings are derived via corresponding encoders. Nonetheless, all embeddings should use the same embedding space in order to facilitate the search. In our work, we use CLIP embedding space to generate embeddings. Each image in the dataset is embedded using the CLIP vision encoder (CLIP-VE), and each piece of the text is embedded using the CLIP text encoder (CLIP-TE). The same goes for the query. Then, the cosine similarity is found between the embedding of the query and the embeddings in the vector database, irrespective of its original data type. 

\begin{figure}[t]
\centerline{\includegraphics[width=0.9\linewidth]{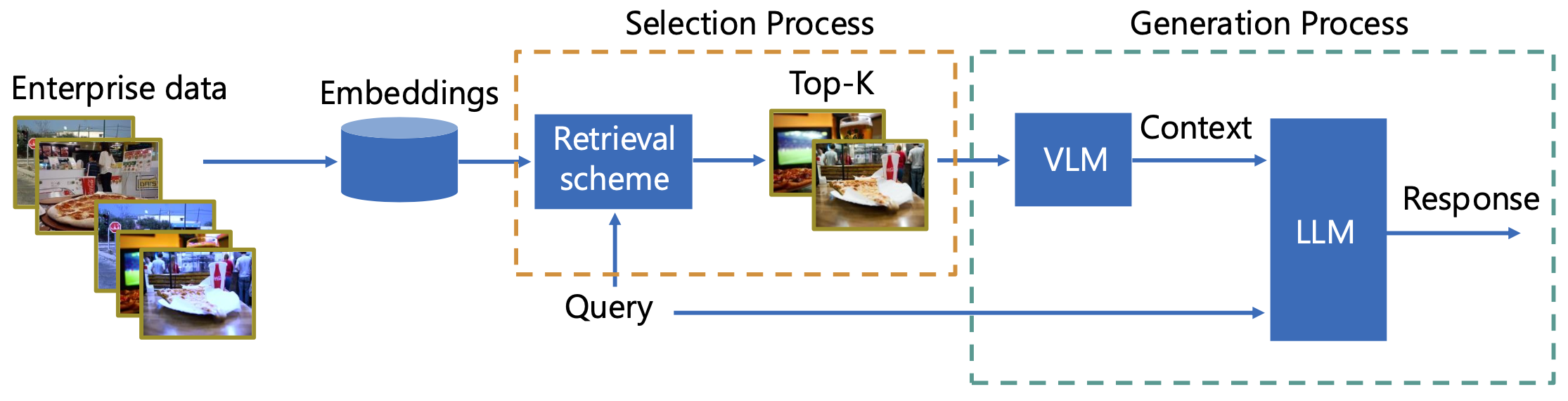}}
  \caption{RAG scheme structure.}
\centering
\label{fig:RAG}
\vspace*{-0.5cm}
\end{figure}

Based on such preprocessing, a vector database is produced that is composed of a pair of embeddings and the reference to each piece of information (original data) in the enterprise database. The RAG system is also composed of several blocks which are used during the operation. A selection block retrieves the relevant pieces of the enterprise data by performing a cosine similarity search using the embeddings for each piece of the data. 
The final stage of the selection process is determining the \emph{raw context} which is composed of the corresponding original pieces of the data for the top-$k$ entries in terms of the cosine similarity.
The last block of the RAG system is composed of a generation block which takes the retrieved context and generates a RAG response based on the query. In multi-modal RAG, this block may be a MLLM (such as GPT4o), capable of directly handling multi-modal data. Alternatively, this block can consist of engines that generate a text-based context for each piece of retrieved data, which is then collectively used as input to an LLM (such as LLAMA or GPT3.5) to generate the response. For instance, if the original data is in text form, it is used as is, whereas a VLM is employed to convert any images into text descriptions. Both ``selection block" and ``generation block" are illustrated in Fig.~\ref{fig:RAG}.

\section{RAG-Check}

\begin{figure}[!h]
\centerline{\includegraphics[width=1\linewidth]{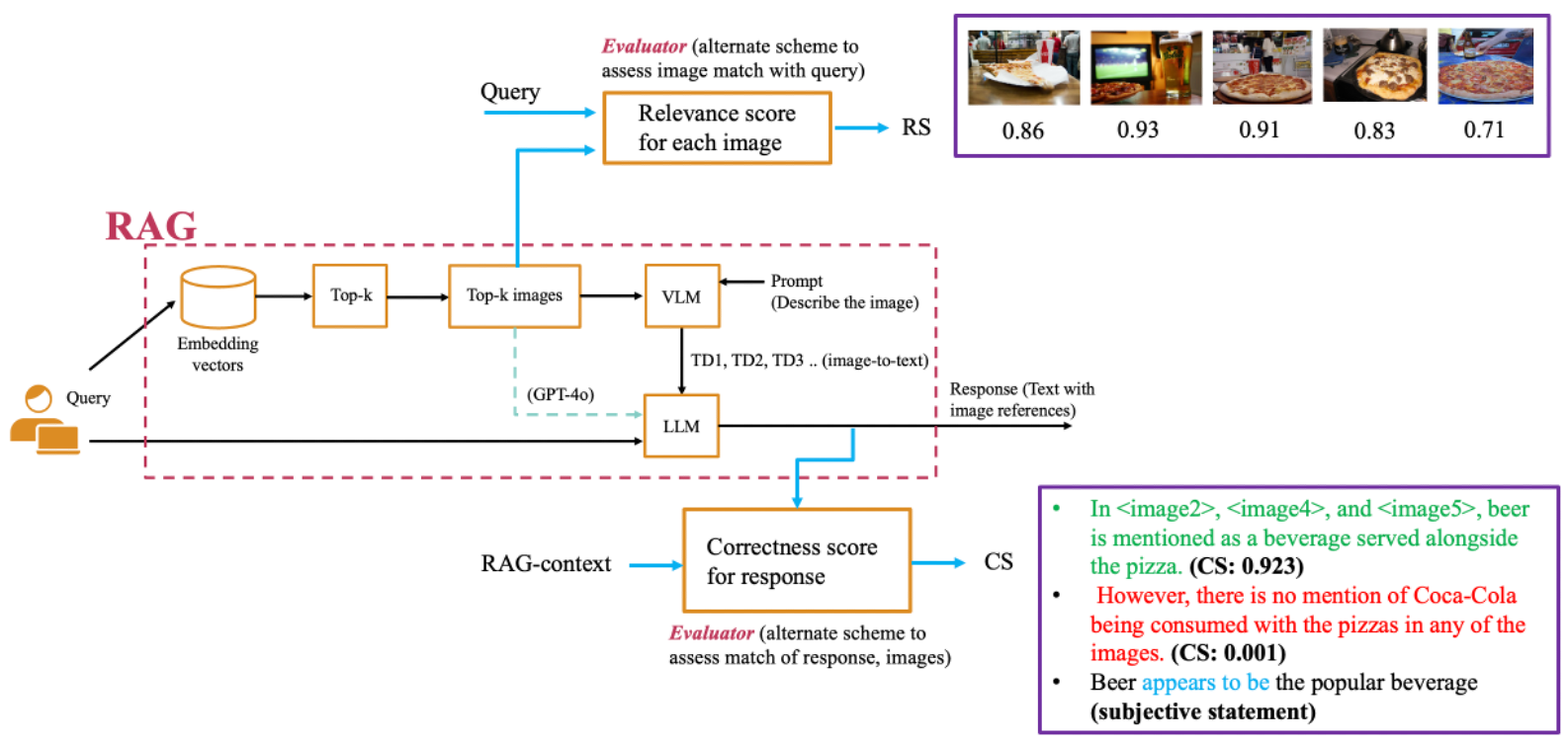}}
  \caption{Overlay of RAG-check structure on RAG block diagram.}
\centering
\label{fig:RAG-check}
\vspace*{-0.2cm}
\end{figure}

Our main objective in RAG-check is to evaluate both the selection process (i.e., how relevant the ``retrieved documents" are to the query) and the generation process (i.e., which parts of the LLM response are supported by the RAG context). Note that the generation process comprises response generation and possibly text-based context generation. In particular, we introduce two quantitative measures: the relevance score (RS) and the correctness score (CS) for multi-modal RAG schemes to determine the performance of the selection and generation process, respectively. The RS provides a quantitative measure between zero (least relevant) and one (most relevant) for each retrieved piece of information. 
In order to assess the quality of the generated output in more details, we break down the generated response by RAG into smaller pieces, namely \emph{spans}, and calculate the CS for each span. Therefore, our RAG-check model consists of three main blocks: 1) Splitting the RAG-generated response into spans and categorizing each span as either a subjective or factual statement. 2) RS model. 3) CS model. Fig.~\ref{fig:RAG-check} shows the 
RAG-check model overlaid on the RAG system model. Fig.~\ref{fig:RAG-check} also illustrates an example scenario where the query ``Is beer more popular to drink with pizza or Coca-Cola?" is submitted to a RAG system, and the RAG selects the 5 most relevant images from the enterprise dataset and produces a response containing 3 statements. The RS block calculates the relevancy score for each image shown in Fig.~\ref{fig:RAG-check}. Also, the generated response is broken into 3 spans where one is ``subjective" and the other two are marked with the respective correctness score calculated by CS block. In the following, we further explain each of these three blocks of the RAG-check system.

\subsection{Partition and Categorization} \label{Sec:partition_catagorization}
Depending on the type of span that we define (phrase, atomic statement, sentence, or paragraph), we split the generated response of the RAG scheme into those spans. We define \emph{atomic statements} as full sentences that are self-sufficient in expressing a meaning without the need to be evaluated along with another sentence or part of the original text. For example, consider the response as: ``In the image, the desk is red and shiny. It is made of wood that is decorated with nice inlays." In breaking down the second sentence in the response, the personal pronoun ``It" has to be replaced with ``The desk" to make this statement self-sufficient without the need to be evaluated with the first statement. We use GPT3.5 with a proper prompt to partition the response to atomic statements. The prompt is engineered in a way that whenever it is needed, the personal pronouns (such as `she', `he', `it', and `they'), demonstrative pronouns (such as `this' and `that'), progressive pronouns (such as `his'), etc. are replaced with the proper reference from the original text. 

Next, we identify whether the given span is a subjective statement or an objective statement. A subjective statement is one that may depend on human feelings, or is subject to personal viewpoints, experiences, or perspectives. Subjective statements are not scorable and are usually difficult to infer directly from the image or they are debatable.
We perform an algorithm to label the subjective statements. In particular, we search for modal verbs (e.g., ``could", ``might"), opinion indicators (e.g., ``believe", ``feel"), hedging phrases (e.g., ``it seems,"), uncertain quantifiers (e.g., ``some", ``many"), adverbs of frequency and degree (e.g., ``often", ``usually"), judgmental adjectives (e.g., ``important", ``useful"), conjectures (e.g., ``it is possible that"), and comparisons or preferences (e.g., ``better", ``prefer") in our algorithm.
Note that GPT3.5 may also be used for categorization of the spans. Nonetheless, we build our system based on our  custom-tailored algorithm since it had better match with human labeling of the statements as ``objective" and ``subjective" based on our evaluation. 

\subsection{Relevance Score (RS) Model}
As defined RS provides a measure of relevance between the query and each piece of the raw context from the enterprise data. The relevancy in RAG systems is usually measured through cosine similarity between the embeddings.
In this work, we address multi-modal RAG by focusing on the similarity between different data modalities. Specifically, we focus on text-based queries and enterprise data which consists of images. We aim to obtain the RS between the query and each of the raw context (image) from the enterprise database separately. Therefore, we design and train a model that receives an image and query as an input and returns a score as a measure of the relevancy between an image and the query.

\begin{figure}[!h]
\centerline{\includegraphics[width=1\linewidth]{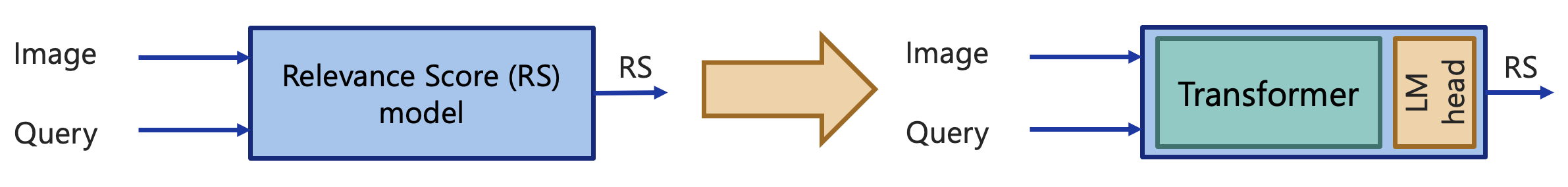}}
  \caption{General structure to quantify relevancy of image to query.}
\centering
\label{fig:optimal_way}
\end{figure}

Even though the relevancy is measured by cosine similarity between the embeddings in RAG, this approach is arguably not optimal. Based on the current state-of-the-art, relevancy can be better examined by performing cross-attention between the query and the image in a transformer as in Fig.~\ref{fig:optimal_way}. Successive cross-attention units in the transformer integrate the query with the information from the image patches, and through training, such structure can outperform the use of cosine similarity between the embeddings of the query and image. Nonetheless, this approach incurs significant complexity, e.g., the calculation of RS in our design using the same GPU machine is 35 times slower than the computation of the cosine similarity. This is the primary reason that the RAG system relies on cosine similarity to search the enterprise data (through stored embedding in a vector database that is produced in a preprocessing stage).

Once the transformer module is designed properly, its output contains valuable information about the relevancy between the query and the image. At this point, a neural network head is trained to extract this information out of the produced embedding by the cross-attention module in the form of a single real number between zero (representing no relevance) and one (representing complete relevance). 

\begin{figure}[!h]
\centerline{\includegraphics[width=1\linewidth]{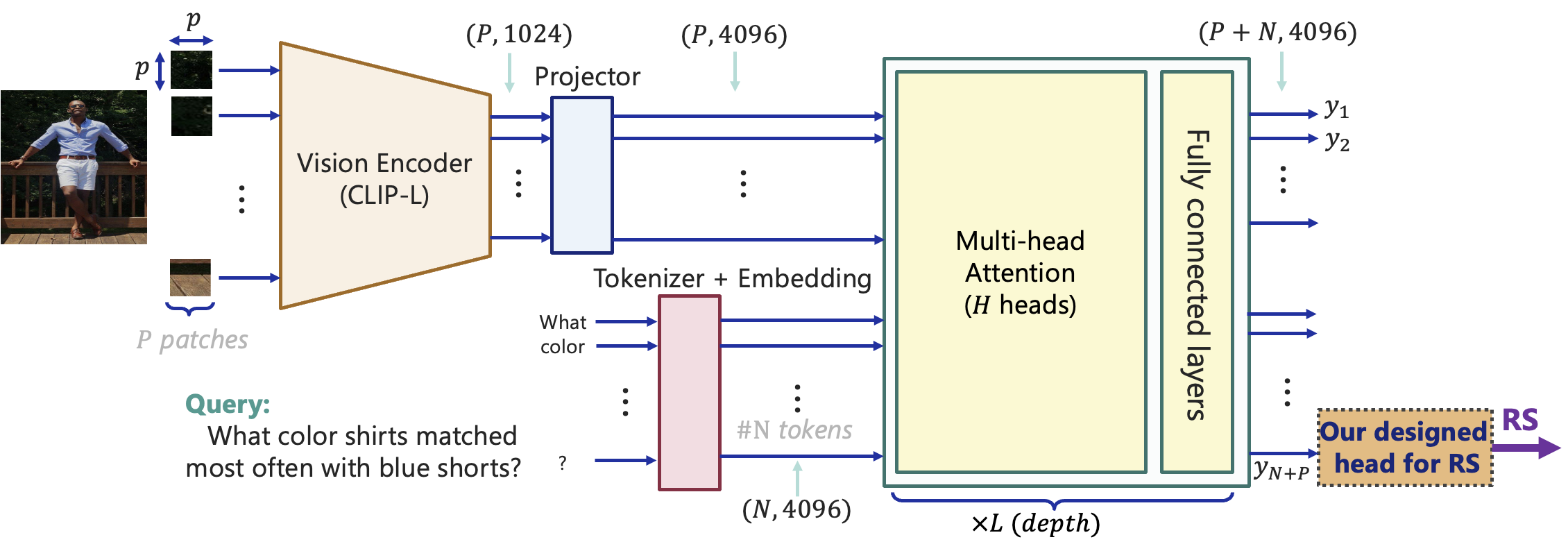}}
  \caption{RS model structure.}
\centering
\label{fig:RS_model}
\end{figure}

Hence, to train for RS, we propose the model as shown in Fig.~\ref{fig:RS_model}. This model is composed of 5 blocks discussed in the following in more details. 
(i) Vision encoder, which encodes patches of images separately. 
We use CLIP large as a vision encoder which has a transformer architecture to encode image patches. 
(ii) Projector: 
The embeddings of patches obtained from the vision encoder need to be translated to a language that is known by the transformer block, and the projector performs this conversion between the output embeddings created by the vision encoder and the input embeddings to the transformer.
(iii) Tokenization and embedding: Each word in the text query has a corresponding token that will be mapped to an embedding. Some tokens require special treatment. For example, if a user query contains a special image token ($<$Image$>$), the token will be replaced with the embeddings of the patches. If there is no reference to an image, the system will add image embeddings to the beginning of user query embeddings. As a result, we have $N+P$ embeddings in total where $N$ is the number of user query tokens (with the exception of special tokens) and $P$ is the total number of patches. The embedding space is $d$ dimensional. 
(iv) Transformer block: The entire $N+P$ embeddings are processed by the transformer, which contains $L$ transformer blocks each containing multi-head attention (MHA) unit with $H$ attention heads and a fully-connected layers. The attention mechanism in the transformer is used to find the relation between different patches of image and user query. The output of the transformer unit is a vector of $N+P$ embeddings each with dimension $d$. 
(v) LM head: The last generated token by the transformer, i.e., $y_{N+P}$ in Fig. \ref{fig:RS_model}, is an embedding of size $d$ which is given to the LM head as input. The LM head is a fully connected layer that maps dimension $d$ to 1 which is trained to represent RS.

Training the entire model from scratch to learn both language and the relationship between language and images requires vast amounts of data and computational power. Therefore, for the backbone of our system, we leverage the weights from the current state-of-the-art model. Specifically, we use LLaVA \cite{liu2024visual} weights, which include the clip-vit-large-patch14-336p as the vision encoder and LLAMA \cite{touvron2023llama} as the LLM decoder, to combine the image patches and query tokens. We modify the final head of LLaVA, which originally converts embeddings into vocabulary, and replace it with a dedicated head that maps the LLM embedding dimension $d$ to a 1 (single output).
To train the RS model (specifically our head), we use a training dataset consisting of triplets $(\mathcal{I}, s_p, s_n)$, where $\mathcal{I}$ is the image, $s_p$ is a positive statement about the image $\mathcal{I}$, and $s_n$ is a negative statement about the image $\mathcal{I}$. We define RS model as $\mathcal{M}$. The output of our model with the given statement $s$ is a vector $y = M(\mathcal{I},s)$ of dimension $N+P$. For the sake of short notation, we use $y_{-1}(\mathcal{I},s)$ to represent the last embedding output of the LLM decoder (in Fig.~\ref{fig:RS_model}) given an image and a statement. 
We use the following template in processing the user query {$s$} by the RS model: ``Evaluate the relevancy of the given statement with the image $<$image$>$. Evaluate by either `relevant' or `irrelevant'. The statement is: {$\{s\}$}."

We use a modified version of the reinforcement learning with human feedback (RLHF) loss function \cite{ziegler2019fine} to train our RS model. In the original RLHF model, even though we have data indicating both highly preferable and less preferable instances, the loss function only ensures that the highly preferable instance receives a higher score than the less preferable one and there is no lower or upper bound of the loss function. However, since we want to assign a score that falls within the range of [0, 1] for any given statement and image, we modify the RLHF loss function as follows:
\begin{equation}
    \mathcal{L} = -\log \left(\sigma\left(y_{-1}(\mathcal{I},s_p)\right)-\sigma\left(y_{-1}(\mathcal{I},s_n)\right)\right),
\end{equation}
where $\sigma$ is our softmax operator. During the inference, given a pair of $(\mathcal{I}, q)$, we can obtain the RS as:  
\begin{equation}
    \text{RS} = \sigma\left(y_{-1}(\mathcal{I},q)\right).
\end{equation}


\subsection{Correctness Score (CS) Model}
When the RAG response is generated, we apply the partition mechanism above
to break the entire generated response $r$ into the spans, in here atomic statements $\{s_i\}_{i-1}^L$ where $L$ is the number of spans. For each of the atomic statements, we use the categorization algorithms mentioned above
and mark the statements as subjective or objective. For objective statements, we use CS models to obtain its correctness score. For each atomic statement, the CS model has access to all retrieved images along with the statement for the correctness measure evaluation. 

In terms of structure, the difference between the CS model and the RS model is its ability to work with multiple images rather than a single image. Hence, we exploit VILA instead of the LLaVA model and to simplify training, we adopt the weight from the VILA model in Fig. \ref{fig:RS_model} for the CS model. The reason for this is that LLaVA is not originally trained on multiple images, which limits its performance when it makes inferences on multiple images. This in turn affects the performance of the CS score which is the fine-tuned model with a dedicated LM head. In contrast, VILA has a similar structure to the LLaVA model but is trained specifically for multi-image inference. 

The training process for the dedicated LM head in our CS model is similar to that of the RS model. The template that we add to the beginning of each statement $s_i$ is as follows: ``I am giving you $k$ images. Evaluate this statement with these images and answer by either `correct' or `incorrect': {$\{s_i\}$}". When there is no reference to a particular piece of context, CS is found between the statement and the entire pieces in the raw context. However, in the calculation of CS for a statement that has particular references to pieces of the context, CS is found between the statement and only the referred pieces of the context in the statement. For example, if an evaluation of the statement: $s_k = $ ``A boy with a cowboy hat is riding a white horse in $<$image1$>$", the CS is only computed by using the template: ``I am giving you a statement. Evaluate this statement and answer by either `correct' or `incorrect': {$s_k$}", where the embeddings of image1 are inserted in the position of the token $<$image1$>$. 

\section{Numerical Result}\label{sec:numerical-results}

In this section, we present numerical results on the training and evaluation of our proposed RS and CS models.
The intended use case of the RS and CS score in this paper is to evaluate any particular instances of response invoked by a query when using a given RAG. In this process, RS assesses the performance of the selection scheme of the RAG and CS assesses the performance of the generation scheme (response and possibly combined with context generation) of the RAG.
In the following, we first discuss the training database, the specifics of the model and hyper-parameters, and then provide the evaluation results for the RS and CS models. We also evaluate the matching and alignment of our proposed scores with the data from human evaluation. We finally use our score to compare the average performance of different RAG schemes.

\subsection{Dataset Specification}
In this subsection, we describe the dataset used for training, validation, and evaluation of the RS and CS models. We create and utilize a dataset comprising 121,000 samples partitioned randomly for training (101K samples), validation (10K samples) and evaluation (10k samples). This dataset is a mixture of two sources: a ChatGPT-derived database and a database containing human evaluator samples \cite{gunjal2024detecting}. The human evaluator database consists of image-statement pairs from the COCO image dataset \cite{lin2014microsoft}, where the statements were marked as either positive or negative by human evaluators. Additionally, we task GPT-4o with generating positive and negative statements for an additional 2,000 COCO images. The statements are then checked by human for accuracy of generation. Since we aim to use the RLHF loss function for training, These two datasets (GPT derived and the datasets in \cite{gunjal2024detecting}) are then combined such that each sample in the final database contains a triplet: an image, a positive statement, and a negative statement. An example of the dataset is given in the following.
\begin{figure}[!h]
\centerline{\includegraphics[width=0.8\linewidth]{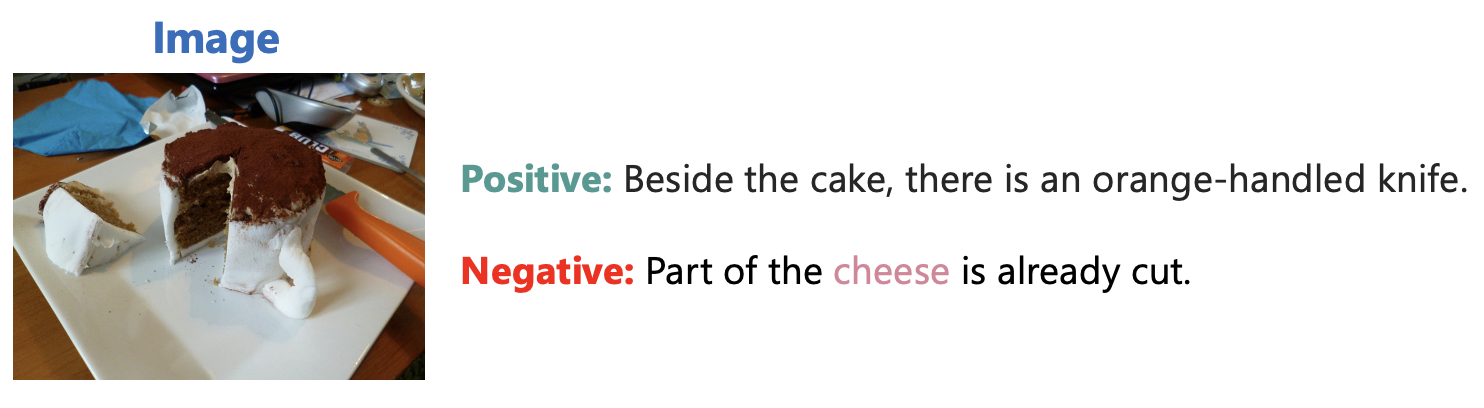}}
  \caption{A sample entry of dataset used for training/evaluation.}
\centering
\label{fig:Example_dataset}
\vspace*{-0.2cm}
\end{figure}

To evaluate the entire RAG system and measure its alignment with human evaluators, we created two datasets (described 
above) by collecting human preferences regarding both RAG selection and generation processes.

\subsection{Model and Hyperparameters}
For the structure of the RS model, we use ``clip-vit-large-patch14-336" \cite{radford2021learning} as a vision encoder, and a projector to map the vision encoder output to proper embedding. We use llama-1.5v \cite{touvron2023llama} as a language decoder model and replace the language model final layer head with our custom head which maps from dimension 4096 to 1. To benefit from the pre-training, we use the trained weight of the LLaVA model for the vision encoder projector and language model. We train our RS model on GPU A100. For the CS model, the vision encoder is Siglip \cite{zhai2023sigmoid} and the language model is llama-3. Similarly, the last layer of the llama-3 model is replaced with our custom head which maps a layer of size 4096 to 1 and the weights are taken from the VILA model. The learning rate for both training CS and RS models are $\alpha=10^{-4}$.

\subsection{Relevance Score (RS) Performance} \label{sec:RS-LLaVA}
\begin{figure}[t]
\centerline{\includegraphics[width=0.75\linewidth]{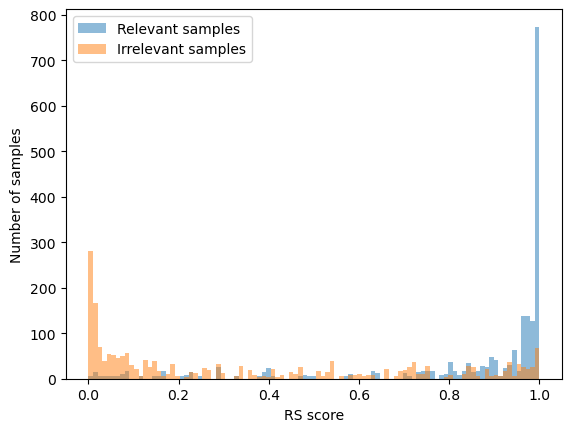}}
  \caption{RS model performance on test data.}
\centering
\label{fig:RS_histogram}
\vspace*{-0.2cm}
\end{figure}

To evaluate our RS score, we use test data to determine how accurately the RS model can detect samples with relevant text to a given image. Fig.~\ref{fig:RS_histogram} shows the score histogram of 2,000 test samples for either of positive and negative statements. The RS score is a number in interval $[0,1]$ and it is not straightforward to evaluate it with respect to an image and query since the dataset entries are only labeled as `relevant' or `irrelevant'. Instead, we can form a \emph{RS-labeler} by using a threshold $0 \leq \eta \leq 1$ to make a hard decision as `relevant/irrelevant' in order to evaluate its performance based on the evaluation dataset entries. One may also use the original LLaVA model with the same prompt used by RS model to form \emph{LLaVA-labeler}.

Table~\ref{TAB:tab1} displays the values for true detection (TD) of relevant samples (denoted as true0) and irrelevant samples (denoted as true1), as well as overall accuracy of the labeling for both RS-labeler with optimized threshold $\eta = 0.7$ and LLaVA-labeler. Table \ref{TAB:tab1} indicates that our model compared to the original LLaVA has greater confidence in determining the relevance of the image to the query as the probability of true detection for both relevant (true0) and irrelevant (true1) samples and hence the overall accuracy for RS-labeler is higher. 



\begin{table}[h]
 \caption{Performance evaluation of scoring models.}
 \label{TAB:tab1}
 \centering
 \renewcommand\footnoterule{\kern -1ex}
 \renewcommand{\arraystretch}{1.3}
 \begin{tabular}{l c  c  c}
 \hline
 Model & Accuracy & true0 & true1\\
 \hline
 LLaVA & 0.724 & 0.695 & 0.746\\
 RS model & 0.865 & 0.909 & 0.831\\ 
 \hline\hline
 VILA & 0.732 & 0.710 & 0.734\\
 CS model & 0.875 & 0.940 & 0.806\\ 
 \hline
 \end{tabular}

 \end{table}

\begin{figure}[!h]
\centerline{\includegraphics[width=0.75\linewidth]{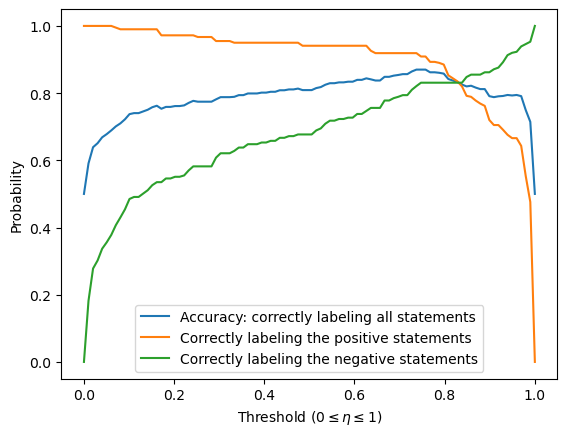}}
  \caption{RS model trade-off performance in changing threshold.}
\centering
\label{fig:RS_tradeoff}
\vspace*{-0.2cm}
\end{figure}

Fig.~\ref{fig:RS_tradeoff} illustrates the effect of changing the threshold on true0, true1, and accuracy. As shown, there is a trade-off: for example, if the threshold is decreased to near 0, almost all samples are detected as relevant, increasing the probability of correctly identifying relevant samples at the expense of incorrectly labeling irrelevant samples. We optimize the threshold $\eta$ where the probabilities of detecting each type of sample are equal to balance the detection of both relevant and irrelevant samples. This threshold may be fine-tuned after training the RS model. 

\subsection{Correctness Score (CS) Performance}
Using similar approach as 
before, we define \emph{CS-labeler} using CS score with threshold $\eta = 0.7$ and \emph{VILA-labeler} to evaluate the performance of CS score based on the evaluation dataset.  Table~\ref{TAB:tab1} shows the evaluation result of CS-labeler and VILA-labeler in detecting the positive and negative statements as correct (denoted as true0) and incorrect (denoted as true1) depending to the given image, respectively. 
Similar conclusions hold for CS score as Table \ref{TAB:tab1} indicates that CS model has greater confidence in detecting the correctness of the statement with respect to the given image in comparison to the original VILA as the probability of true detection for both correct (true0) and incorrect (true1) samples and hence the overall accuracy for CS-labeler is higher. 


\subsection{RAG Scores vs Human Evaluation}\label{sec:RAG-score}
In this section, we aim to show the alignment between RS and CS with human evaluators. For RS, we randomly select 1,000 questions from the test set of the COCO-QA \cite{ren2015exploring} dataset. We also gather 1,281 random images from the COCO validation set and use the CLIP-ViT-large-patch14-334p model to encode both the images and the selected questions. We obtain the top-5 images for each question that had the highest cosine similarity with the encoded query data.
We collect human evaluators' opinions on the relevance of each retrieved image for each question. Evaluators could choose from five options: unsure, no relevance, mild relevance, high relevance, and complete relevance corresponding to a rating 0,1, $\ldots,$ 4, respectively. For each question and retrieved image pair, we calculate the relevancy score with different relevancy scoring algorithms including RS, clip--vit-large-patch14-336, clip-vit-base-patch16, clip-vit-base-patch32, blip \cite{li2022blip}. The following algorithm was used to assess how well the human evaluators' judgments match with relevance scoring measures: 
We consider all possible combinations among the five retrieved images for each question. For each combination, if any image is marked as `unsure' by the human evaluator, we disregard the sample. In cases where there is a difference in human evaluator's opinion between two images, we evaluate how closely the score aligns with this difference.

Specifically, for two nonzero ratings $r_1$ and $r_2$, where $r_1 < r_2$, given two images for a single query, if the score corresponding to the second image is higher, we consider this a match and assign a reward $r_2 - r_1$. Otherwise, it is considered unmatched, and the reward is zero. We then calculate the average reward across all pairs of images for each question and across all questions. 
Table \ref{Tab:RS_match_HE} shows the average reward for different scores including RS, clip-vit-large-patch14, clip-vit-base-patch16, clip-vit-large-patch32, and BLIP. The result indicates that RS has more than 20\% improvement in average reward in comparison to the other scores used for relevancy and has a good match with human evaluator's opinion. 

We use similar methodology to evaluate CS. We use the same test set of 1,000 questions. We fix the selection scheme by using the CLIP-ViT-large-patch14-336 as the vision encoder and cosine similarity to select the top-5 relevant images. Next, we employ LLaVA as the VLM and llama-1.5v as the LLM. The VLM extracts text descriptions from the selected images, while the LLM generates the final response. To evaluate the responses, we partition the generated response into atomic statements using our partitioning algorithm. A human evaluator then assesses each atomic statement for each question, choosing one of three options: correct, incorrect, or subjective. To measure the alignment between our proposed CS and human judgment, we calculate the overlap between our CS and human evaluation. The results show a 91\% match between the CS and the human evaluator's responses.

\begin{table}[h]
\caption{Alignment of RS scores with human evaluators.}
\label{Tab:RS_match_HE}
\centering
\begin{tabular}{@{}lc@{}} 
 \hline
 Relevance Scoring Method  & Value \\ [0.5ex] 
 \hline
 Proposed Relevance Score (RS) & \textbf{0.879} \\ 
 CLIP-vit-large-patch14-336  & 0.689 \\ 
 CLIP-vit-base-patch32  & 0.620 \\
 CLIP-vit-base-patch16  & 0.611 \\
 BLIP-large & 0.491 \\
 \hline
\end{tabular}
\vspace*{-0.2cm}
\end{table}

\subsection{Evaluation of Different RAG Schemes}
In this section, we use our proposed RS and CS scores to evaluate the performance of different RAG schemes in terms of the relevancy of retrieved entries and the correctness of the generated responses.
\subsubsection{RAG Selection Performance:}
\begin{figure}[!h]
\centerline{\includegraphics[width=0.75\linewidth]{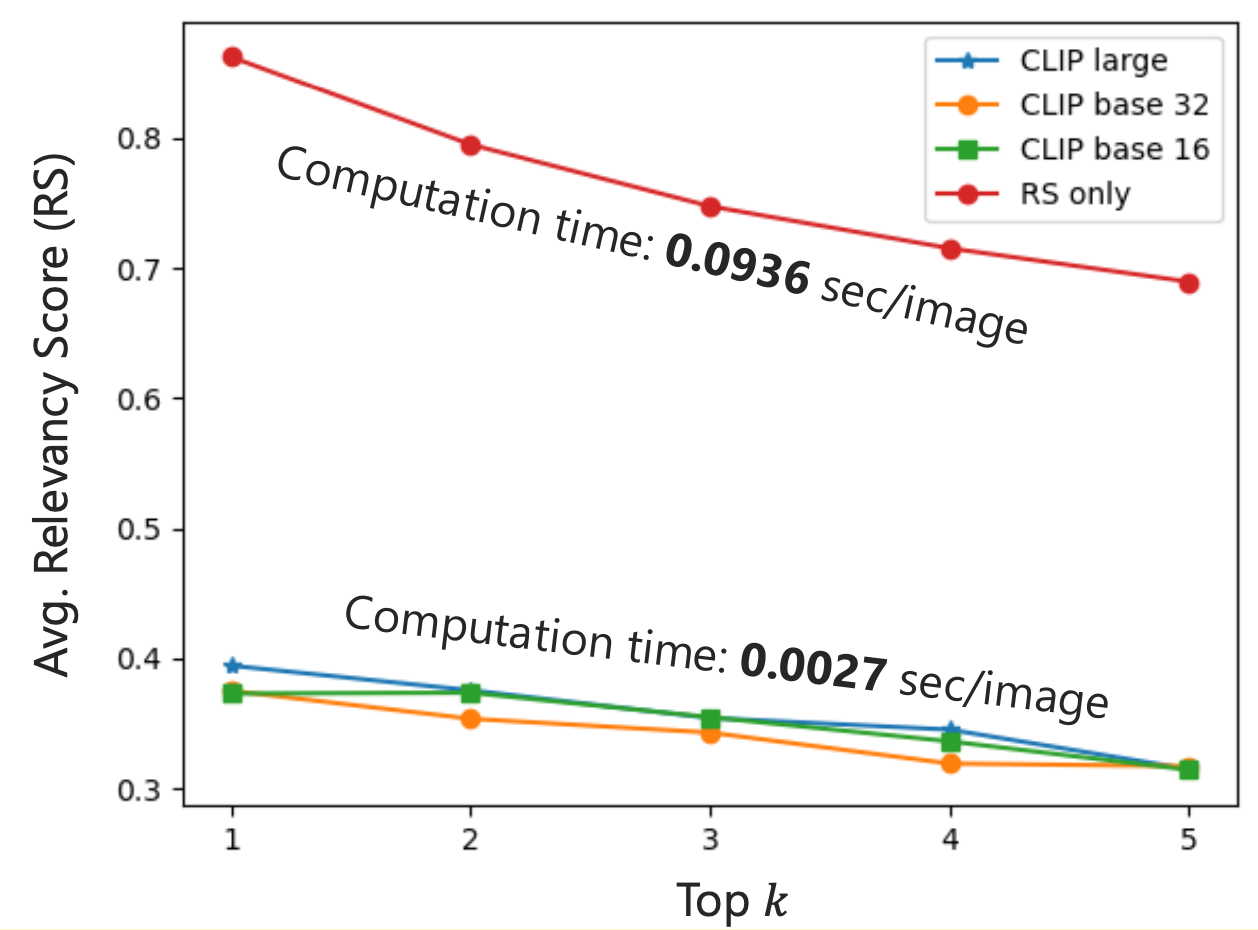}}
  \caption{Comparison of different RAG selection mechanisms with RS model.}
\centering
\label{fig:RS_RAG_compare}
\vspace*{-0.3cm}
\end{figure}

We compare the performance of different selection mechanisms in multi-modal RAG systems. Fig.~\ref{fig:RS_RAG_compare} illustrates the average relevancy score for each of the top-5 retrieved images across 1,000 test questions. As shown, when a RAG system uses CLIP models (clip-vit-large-patch14-336, clip-vit-base-patch32, clip-vit-base-patch16) as the vision encoder to process the images and queries, and cosine similarity to select the top-5 images, the average relevancy of the retrieved images ranges from 41\% to 30\%. In contrast, when using the RS model to calculate the RS score for all possible query-image pairs before selecting the top-5 images, the average relevancy scores significantly improve, ranging from 89.5\% to 71\%. However, using the RS model to score all possible pairs results in a 35× increase in computation compared to the CLIP dot product, even when using a GPU (A100).
The results show a significant improvement in RAG selection performance compared to using CLIP alone, demonstrating how the RS model can enhance the RAG system beyond simply evaluating its performance.

\subsubsection{RAG Context Generation Performance:}
\begin{figure}[!h]
\centerline{\includegraphics[width=0.8\linewidth]{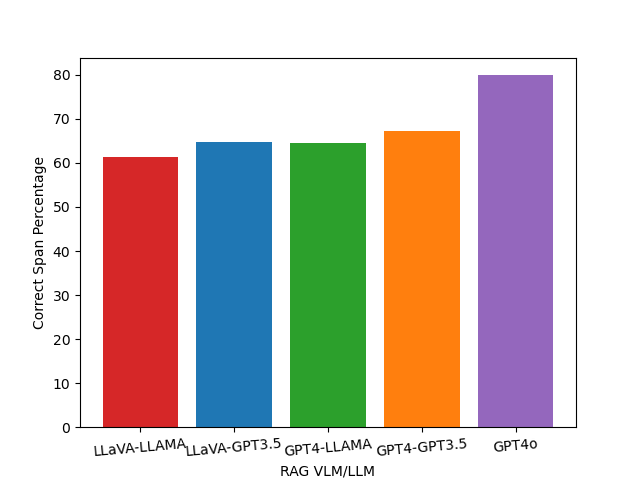}}
  \caption{Comparison of different RAG context and generation mechanism with CS model.}
\centering
\label{fig:CS_RAG_compare}
\vspace*{-0.2cm}
\end{figure}

We use our proposed CS to compare various RAG schemes that incorporate different VLMs and LLMs, or in general different MLLMs. For the VLM component, one can select LLaVA or GPT-4, and for the LLM, options include LLAMA or GPT-3.5. Alternatively, one can combine both VLM and LLM into a single MLLM, such as GPT-4o, to directly generate responses from retrieved images. Fig.~\ref{fig:CS_RAG_compare} illustrates the comparison between the five RAG configurations in terms of context and generation error. The results show that GPT-4o outperforms the other schemes by approximately 20\%, while the remaining RAG schemes exhibit performance within the 60-68\% range.

\bibliographystyle{unsrt}
\bibliography{reference}
\end{document}